\documentclass[11pt]{article}

\pdfoutput=1

\usepackage{eacl2017}
\usepackage{times}
\usepackage{url}
\usepackage{latexsym}
\usepackage{subfig}
\newcommand{\COMMENT}[1] {}

\usepackage{array}
\newcolumntype{C}[1]{>{\centering\let\newline\\\arraybackslash\hspace{0pt}}m{#1}}

\usepackage[vlined]{algorithm2e}

\usepackage{graphicx}

\newcommand{\mmt}{MMT} 

\eaclfinalcopy 

\title{Unsupervised Clustering of Commercial Domains \\for Adaptive Machine Translation}

\author{Mauro Cettolo\\
Fondazione B. Kessler \\
  Trento, Italy \\
  {\tt cettolo@fbk.eu} \\\And
  Mara Chinea Rios\\
  Universitat Polit\`ecnica de Val\`encia\\
  Valencia, Spain\\
  {\tt mchinea@iti.upv.es} \\\And
  Roldano Cattoni\\
Fondazione B. Kessler \\
  Trento, Italy \\
  {\tt cattoni@fbk.eu}   \\}

\date{}

\begin{document}
\maketitle
\begin{abstract}

In this paper, we report on domain clustering in the ambit of an
adaptive MT architecture. A standard bottom-up hierarchical clustering
algorithm has been instantiated with five different distances, which
have been compared, on an MT benchmark built on 40 commercial domains,
in terms of dendrograms, intrinsic and extrinsic evaluations. The
main outcome is that the most expensive distance is also the only one able
to allow the MT engine to guarantee good performance even with few,
but highly populated clusters of domains.

\end{abstract}

\section{Introduction}

In translation industry, a new frontier for MT is to implement generic
systems that adapt on-the-fly to any context. Once fuelled with enough
data and bootstrapped, they work without any further training
iterations and translates sentences in a domain sensitive way, without
any need of prior identification of or adaptation to the domain.

The MT architecture we are working on does not rely on the current
paradigm of handling and using generic and domain-specific training
data in a machine translation system. It assumes that domain
information is not defined a priori or attached to the data, but
instead that any subset of data collected and ingested by the system
might become relevant at some point during the use of the
system. Domain relevance is based on matching the input sentence to
translate, together with some of its context, against all available
data. The result is a probability distribution over the domains, which
is used to activate, with proper weights, the underlying domain
specific models.

Even holding the prompt reaction constraint guaranteed to users by our
architecture, any background processing is allowed in order to further
boost the system behaviour. In fact, among the others, two relevant
issues are still open.  One is the scalability over the number of
``domains'' that can be efficiently handled; the number can grow very
fast since for us a ``domain'' is the specific data provided by each
customer. Another is related to those customers who provide not
enough training data for bootstrapping reliable models.  In order to
keep manageable the cardinality of domains and, if legally possible,
exploit at best all the data previously ingested by the system, a
natural choice is to aggregate similar domains.

In this paper, we report on domain clustering in the ambit of the MT
architecture sketched above. We will provide an empirical positive
answer to the two questions induced by the just mentioned issues:
(i)~is clustering able to aggregate so many domains that with just few
cluster-specific models the MT quality remains adequate?  (ii)~is
clustering able to aggregate domains with few training data such that
the overall performance improves?

After the overviews on domain adaptation
literature (Section~\ref{sec:related}) and on consolidated scientific
knowledge on data clustering (Section~\ref{sec:clstAlg}), the
hierarchical agglomerative clustering actually implemented is
described in Section~\ref{sec:HAC}; five different similarity
measures of clusters are proposed in Section~\ref{sec:distances} and
experimentally compared in Section~\ref{sec:exp}, according to the
framework defined in Section~\ref{sec:mmt}. A discussion, the summary
and the list of investigations planned for the future end the paper.

\section{Related Work}
\label{sec:related}

For optimising performance, the models of machine translation systems
are often specialised on specific domains, like legal, information
technology or medicine.

The specialisation is obtained by training models on text from the
specific domain. Specialised texts can be gathered either by exploiting
supervision or through automatic selection from general
texts. Deciding if a sentence belongs to a given domain can be done by
checking how well it is predicted by a domain specific language model
(through the perplexity), by the tf/idf method commonly employed in
information retrieval, or by the cross-entropy difference method
presented in~\cite{Moore:2010:ACL,Axelrod:2011}.

Instead of discarding part of the training data, multiple models can
be trained on the various domains. Multiple domain-specific models can
be loaded in MT engines that receive input from different domains; the
input is then classified and the proper model activated; for example,
in~\cite{Xu:MTSummit:2007} the classification is done using either
language models or information retrieval methods.

Another way to exploit multiple domain-specific models is the
mixture-model approach which combines the various models, properly
weighting each of them~\cite{Foster:07}. Typically, the combination is
realised as a linear or log-linear model and can involve language,
translation and even alignment models.  The interpolation weights can
be estimated off-line on a development set or on the source side of
the whole test set. On-line estimation is also feasible on the current
sentence (or bunch of sentences) to
translate~\cite{finch-sumita:WMT:2008}; in the ambit of CAT and
interactive MT, the availability of user corrections allows a promptly
and really effective adaptation of the weights~\cite{Mathur:2013}.

If proper meta-information is available, a supervised partitioning of
the training data into domains is allowed. Unfortunately, that is a
rather rare case. More commonly, unsupervised clustering is needed, as
in~\cite{Bungum:Context:2015}, where Self-Organizing Map is used to
create auxiliary language models, the most appropriate of which is
selected on-the-fly for each document to translate.

\section{On the Clustering Algorithms}
\label{sec:clstAlg}

In the following, some excerpts from~\cite{Manning:IR:2008} are fused
to provide a brief introduction to clustering algorithms.

Clustering algorithms group a set of documents into subsets or
clusters. The algorithms' goal is to create clusters that are coherent
internally, but clearly different from each other. In other words,
documents within a cluster should be as similar as possible; and
documents in one cluster should be as dissimilar as possible from
documents in other clusters. Clustering is the most common form of
unsupervised learning.  No supervision means that there is no human
expert who has assigned documents to classes.

Clustering algorithms can be flat or hierarchical.  Flat clustering
creates a flat set of clusters without any explicit structure that
would relate clusters to each other.  Hierarchical clustering creates
a hierarchy of clusters.

Flat clustering algorithms are efficient and conceptually simple, but
have a number of drawbacks. In addition to return a flat unstructured
set of clusters, they typically require a pre-specified number of
clusters as input and are nondeterministic.  On the contrary,
hierarchical clustering outputs a hierarchy, a structure that is more
informative than the unstructured set of clusters returned by flat
clustering.  Hierarchical clustering does not require to pre-specify
the number of clusters and most hierarchical algorithms are
deterministic. These advantages of hierarchical clustering come at the
cost of lower efficiency.  The most common hierarchical clustering
algorithms have a complexity that is at least quadratic in the number
of documents compared to the linear complexity of K-means, one
widely used flat clustering algorithm.

Hierarchical clustering algorithms are either top-down or bottom-up.
Top-down clustering requires a method for splitting a cluster and
proceeds by splitting clusters recursively until individual documents
are reached.

Bottom-up algorithms treat each document as a singleton
cluster at the outset and then successively merge (or agglomerate)
pairs of clusters until all clusters have been merged into a single
cluster that contains all documents. Bottom-up hierarchical clustering
is therefore called hierarchical agglomerative clustering (HAC).

HAC algorithms employ a similarity measure for deciding which clusters
to merge; common similarity measures are: single-link, complete-link,
group-average, and centroid similarity.

In single-link clustering, the similarity of two clusters is the
similarity of their most similar members. This single-link merge
criterion is local. Attention is solely paid to the area where the two
clusters come closest to each other.  Other, more distant parts of the
cluster and the clusters' overall structure are not taken into
account.  

In complete-link clustering, the similarity of two clusters
is the similarity of their most dissimilar members. This is equivalent
to choosing the cluster pair whose merge has the smallest
diameter. This complete-link merge criterion is non-local; the entire
structure of the clustering can influence merge decisions.  This
results in a preference for compact clusters with small diameters over
long, straggly clusters, but also causes sensitivity to outliers.

Group-average agglomerative clustering evaluates cluster quality based
on all similarities between documents, thus avoiding the pitfalls of
the single-link and complete-link criteria, which equate cluster
similarity with the similarity of a single pair of
documents. Group-average similarity computes the average similarity of
all pairs of documents, including pairs from the same cluster (but
self-similarities).

In centroid clustering, the similarity of two clusters is defined as
the similarity of their centroids.  Centroid similarity is equivalent
to average similarity of all pairs of documents from different
clusters. Thus, the difference between the group-average similarity
and the centroid similarity is that the former considers all pairs of
documents in computing average pairwise similarity whereas the latter
excludes pairs from the same cluster.

\subsection{Evaluation}
\label{sec:evaluation}

An unsupervised clustering can be evaluated in two ways: intrinsically,
according to properties of the clusters, or extrinsically, according
to the performance on a task which uses the clustering.

For intrinsic evaluation, the Silhouette coefficient\footnote{en.wikipedia.org/wiki/Silhouette\_(clustering)}~\cite{Rousseeuw:1987:JCAM}
can be used, which measures how similar an object is to its own
cluster (cohesion) compared to other clusters (separation).

For each datum {\em i}, let {\em a(i)} be the average dissimilarity of
{\em i} with all other data within the cluster which {\em i} belongs
to.  {\em a(i)} can be interpreted as how well {\em i} is assigned to its
cluster (the smaller the value, the better the assignment). 

The average dissimilarity of {\em i} to a generic cluster {\em c} is
defined as the average distance from {\em i} to all points in {\em
  c}. Let {\em b(i)} be the lowest average dissimilarity of {\em i} to
any other cluster, of which {\em i} is not a member. The cluster with
this lowest average dissimilarity is said to be the ``neighbouring
cluster'' of~{\em i} because it is the next best fit cluster for~{\em
  i}. The Silhouette value for {\em i} is defined as:

$$
s(i) = \frac{b(i) - a(i)}{ max\{a(i), b(i)\} }
$$

\noindent From the definition, it results that:
$$
    - 1 \le s(i) \le 1
$$

For {\em s(i)} to be close to 1 it is required that $a(i)\ll~b(i)$. As
{\em a(i)} is a measure of how dissimilar {\em i} is to its own
cluster, a small value means it is well matched. Furthermore, a large
{\em b(i)} implies that {\em i} is badly matched to its neighbouring
cluster. Thus an {\em s(i)} close to one means that the datum is
appropriately clustered. If {\em s(i)} is close to negative one, then
by the same logic it is seen that {\em i} would be more appropriate if
it was clustered in its neighbouring cluster. An {\em s(i)} near zero
means that the datum is on the border of the two clusters.

Note that when a cluster contains only a single object {\em i}, {\em
  a(i)} cannot be defined; following~\cite{Rousseeuw:1987:JCAM}, we
simply set {\em s(i)} to zero, an arbitrary but neutral choice.

The average {\em s(i)} over all data of a cluster is a measure of how
tightly grouped all the data in the cluster are, while the average
{\em s(i)} over all data of the entire dataset is a measure of how
appropriately the data have been clustered.

\section{The HAC Algorithm}
\label{sec:HAC}

Algorithm~\ref{alg:clstrng} shows the pseudo-code of the hierarchical
agglomerative procedure we have implemented.
Two functions play the main role, namely {\tt d()}, which computes
somehow the ``distance'' between two clusters, and {\tt evaluate()},
which, given a clustering, provides a score of its quality. {\tt Q[]}
is a data structure for storing triples $q=(\delta,i,j)$ where
$\delta$ is the distance between the $i^{th}$ and the $j^{th}$
clusters; if {\tt d()} is really a distance (and hence identity of
indiscernibles and symmetry are satisfied conditions), {\tt Q[]} is
a strict (upper or lower) triangular matrix.

\RestyleAlgo{boxruled}
\begin{algorithm}
\footnotesize
\SetAlgoLined 
\KwData{$D=[D_1 \ldots D_N]$}
\KwResult{$N$ scores, one for each generated clustering}
\For{$i,j=1\ldots N, i<j$}{ compute $\delta={\tt d}(i,j)$\\ store [$\delta,i,j$] in $Q$}
output ${\tt evaluate}(D)$  \hspace{5mm} {\scriptsize // quality score of the initial clustering} \\
\For{$t = 1 \to N-1$}
  {   
    $[\bar{\delta},\bar{i},\bar{j}] = \arg\min_{q=[\delta,i,j]\in Q} q[1]$\\
    remove $D_{\bar{i}}$ and $D_{\bar{j}}$ from $D$\\
    remove entries 
{\scriptsize $[*,\bar{i},*]$,$[*,*,\bar{i}]$,$[*,\bar{j},*]$,$[*,*,\bar{j}] $} 
from $Q$\\
    insert $D_{\bar{i}} = D_{\bar{i}} \circ D_{\bar{j}}$ into $D$\\
    \ForEach{$l: [*,l,*]\in Q$}{
      $\delta={\tt d}(l,\bar{i})  $ \hspace{29mm} {\scriptsize //$ \ \ == {\tt d}(\bar{i},l)$}\\ 
      \eIf{$l<\bar{i}$}{
        store [$\delta,l,\bar{i}$] in $Q$
      }({{\scriptsize \hspace{45mm}// $ \ \ l>\bar{i}$}}){ 
        \small
        store [$\delta,\bar{i},l$] in $Q$
      }
    }
        output ${\tt evaluate}(D)$ \hspace{2mm} {\scriptsize  // quality score of the $t^{th}$ clustering}
}
\caption{the HAC}
\label{alg:clstrng}
\end{algorithm}

The algorithm takes as input the set of $N$ documents to cluster. Each
of them is considered as a single cluster, hence the size of the
initial clustering is $N$. The distances between any pair of
documents are computed and stored in the entries above the main
diagonal of the matrix {\tt Q[]}. Before starting to iterate, the
quality of the initial clustering is output.

At each iteration, the two closest clusters are identified ($\arg\min$
operation performed on $q[1]$, the first component of triples $q$);
their rows and columns in {\tt Q[]} are removed; then, they are merged
($\circ$) and the smallest of their two indexes is assigned to the new
cluster; finally, the distance of the new cluster from any other
cluster is computed and stored in the upper part of {\tt Q[]}. Before
ending the iteration, the quality of the new clustering is output.

The algorithm is suboptimal, since the local decision to
merge the two current closest clusters cannot be
backtracked. Moreover, no stopping criterion is designed, hence at the
end all initial clusters are merged into a single cluster; that allows
to have plots covering the whole range of clusterings in between the
two extremes ($N$ and $1$ clusters), as we will see. On the other
side, an ending rule could be easily devised looking at the value of
$\bar{\delta}$, for example comparing it to some threshold.

\subsection{Tested similarity measures}
\label{sec:distances}

In our context, the single points to be clustered are
translation memories (TMs), that is collections of sentence pairs. The
distance ${\tt d}^{\tt trg}()$ between the target sides of two TMs
$D_i$ and $D_j$ can be measured by means of the cross-perplexity:
  $${\tt d}^{\tt trg}_{\tt PP}(i,j) = PP_{LM^{\tt trg}_{j}}(D^{\tt trg}_{i})\oplus PP_{LM^{\tt trg}_{i}}(D^{\tt trg}_{j})$$
\noindent
where $D^{\tt trg}_x$ is the target text in the TM $D_x$, $LM^{\tt
  trg}_x$ is the language model estimated on $D^{\tt trg}_x$ and
$PP_{LM}(D)$ is the perplexity of $LM$ measured on $D$, which
indicates how well the probability distribution $LM$ predicts the text
$D$. $\oplus$ indicates the proper sum of perplexities.

As seen in Section~\ref{sec:clstAlg}, the distance between clusters
with more than one TM can be computed by measuring the similarity of
two single TMs: the closest in the single-link case, the farthest in the
complete-link.  Hereafter, they will be indicated as ${\tt
  d_{PPcls}}()$ and ${\tt d_{PPfar}}()$, respectively.

Group-average and centroid are instead similarity measures which
involve all the points of clusters. Given the peculiarity of our case,
instead of group-average or centroid, in order to involve all data of
clusters, a natural choice is to really merge TMs: when $D_i$ and $D_j$
are clustered, instead of considering the new cluster as a collection
of two separate TMs, it is though as a new single TM $D_{ij}$ which is
the concatenation of $D_i$ and $D_j$. This way, the distance ${\tt
  d}_{\tt PP}()$ defined above can be computed for any pair of clusters
generated by the agglomerative algorithm.

A variant of ${\tt d}_{\tt PP}()$ consists in looking ahead the impact
of merging a pair of clusters by computing:

\bigskip

${\tt d}^{\tt trg}_{\Delta}(i,j)= PP_{LM^{\tt trg}_{ij}}(D_{ij}) $

\hspace{8mm}$ - \left( PP_{LM^{\tt trg}_{i}}(D^{\tt trg}_{i})\oplus PP_{LM^{\tt trg}_{j}}(D^{\tt trg}_{j}) \right)$

\bigskip

\noindent
that is the difference between the perplexity on $D_{ij}$ of the
LM estimated on it and the cumulative perplexity on $D_i$ and $D_j$
of the two corresponding LMs. The smaller the difference, the
more convenient is to merge the two clusters. ${\tt d}_{\Delta}()$ is
more expensive than ${\tt d}_{\tt PP}()$ because at each iteration it
requires to train and evaluate $LM_{ij}$ for any pair $(i,j)$, while
the latter just for the pair selected by the $\arg\min$ operation.

\smallskip

As described in Section~\ref{sec:mmt}, given a source document, the
module of our system named Context Analyser (CA) generates a domain
distribution vector. It can then be exploited to measure the distance
among clusters by means of:
$${\tt d}^{\tt src}_{\tt CA}(i,j)= 2 - \left( {\Pr}^{\tt src}_{\tt CA(D^{\tt src}_i)}(j) + {\Pr}^{\tt src}_{\tt CA(D^{\tt src}_j)}(i) \right) $$

\noindent
where ${\Pr}_{\tt CA(D)}()$ is the discrete distribution provided by the
CA over the document $D$. The rationale behind ${\tt d_{CA}}()$ is that
the more $D_j$ ($D_i$) suits the context $D_i$ ($D_j$), the higher 
$\Pr()$ and then the lower ${\tt d}_{\tt CA}(i,j)$.

\smallskip
Note that all the above $\tt d()$ are defined over either the target
or the source side, but they hold for the opposite side as
well. Moreover, the computation of the distance ${\tt d}(l,{\bar{i}})$
in Algorithm~\ref{alg:clstrng} involves the training of new LMs in all
the cases but ${\tt d_{PPcls}}()$ and ${\tt d_{PPfar}}()$, for which
the values stored in $Q$ during the initialisation phase can be reused
in any iteration, making those two distances definitely more efficient
than the others.

\section{Working Setup}
\label{sec:mmt}

The {\mmt} project,\footnote{\url{www.modernmt.eu}} described in~\cite{mmt:deliverable1.2,mmt:deliverable3.1},
features an on-line domain adaptation.  A context analyser is
employed whose training consists in the creation of a database built
on the source side of training data alongside the domain provenance
meta-information. At translation time, given a source text window or
an entire document, the context analyser generates a domain distribution vector
including the top matching domains available in the training data.
The vector is passed on to the MT engine that will properly adapt the
translation and language models to the input document: in the former
case, by biasing the sampling of translation pairs in the suffix
array~\cite{Germann:PBML:2015}, while in the latter case, by linearly
combining domain specific language models.

\subsection{Data}

Textual data in {\mmt} is divided into ``domains''. From the
commercial translation service provider's point of view, the most
straightforward manifestation of ``domain'' is the customer-specific
TM: the archive of all documents translated by
the provider for a specific customer. Large translation clients can
use product- or business-area-specific TMs, but it happens that TM
contents are heterogeneous; therefore, the {\mmt} concept of ``domain''
differs from the usual meaning given to the word ``domain''.

Documents were  collected from the two major sources of TMs
available to {\mmt}: the TAUS Data Cloud and Translated's
MyMemory. Details are provided in~\cite{mmt:deliverable5.2}.

For MT evaluation purposes, an English-Italian benchmark was built. It
includes the 30 largest TMs from the MyMemory database; the provenance
of the documents varies from software documentation to legal documents
and advertising. From the TAUS Data Cloud, 33 further TMs were also
added.  This benchmark will be referred to henceforth as Benchmark
1.1.

Data in Benchmark 1.1 was split into training, development and test
sets. In order to evaluate the performance of the translation engine
in a real scenario, the final composition of development and test sets
includes all 30 domains (i.e. translation memories) from MyMemory and
a selection of 10 domains (translation memories) from TAUS Data
Cloud. Table~\ref{tab:resource:bilingual} provides statistics on
bilingual data sets; figures refer to untokenized texts.

\begin{table}[ht]
  \centering
  \begin{tabular}{c|ccc}
    \hline
set   & \#sent & $\mid$src$\mid$ & $\mid$ trg$\mid$ \\
    \hline

train & 5.3M & 87.6M & 82.3M \\
dev   & 1,000 & 15,407 & 15,485 \\
test  & 1,000 & 14,950 & 15,001 \\
    \hline
  \end{tabular}
  \caption{Statistics on bilingual resources.}
  \label{tab:resource:bilingual}
 \end{table}

\section{Experimental Results}
\label{sec:exp}

Different instances of Algorithm~\ref{alg:clstrng}, one for each
distance defined in Section~\ref{sec:distances}, were run to cluster
the 40 TMs of the test set; processing times are reported in
Table~\ref{tab:proctime}. In the following sections, first the
dendrograms visualize how the various instances of the algorithm
perform the clustering step by step; then, intrinsic and extrinsic
evaluations are provided.

\begin{table}[ht]
  \centering
  \begin{tabular}{C{14mm}C{19mm}C{15mm}C{14mm}}
    \hline
${\tt d}^{\tt trg}_{\tt PP}()$ &  ${\tt d^{\tt trg}_{PPcls/far}}()$ & ${\tt d}^{\tt trg}_{\Delta}()$ & ${\tt d}^{\tt src}_{CA}()$\\
    \hline
15 & 10 & 100 & 50\\
    \hline
  \end{tabular}
  \caption{HAC instances processing time (min).}
  \label{tab:proctime}
 \end{table}

\subsection{Dendrograms}

\begin{figure*}[ht]
\centering
 \subfloat[]{\includegraphics[width=0.5\textwidth]{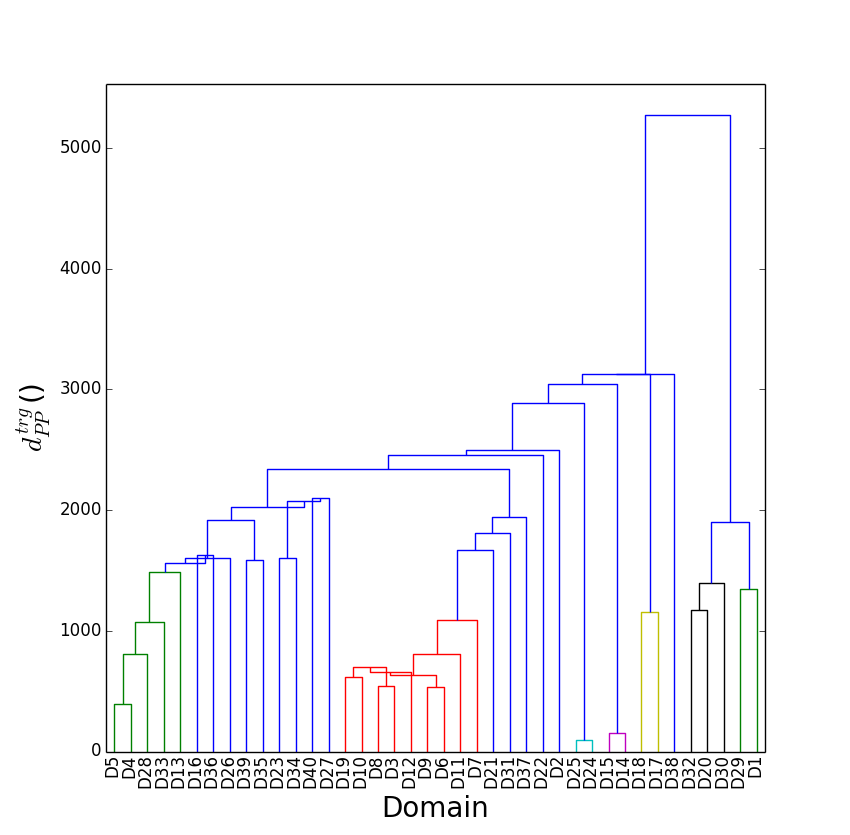}} 
 \subfloat[]{\includegraphics[width=0.5\textwidth]{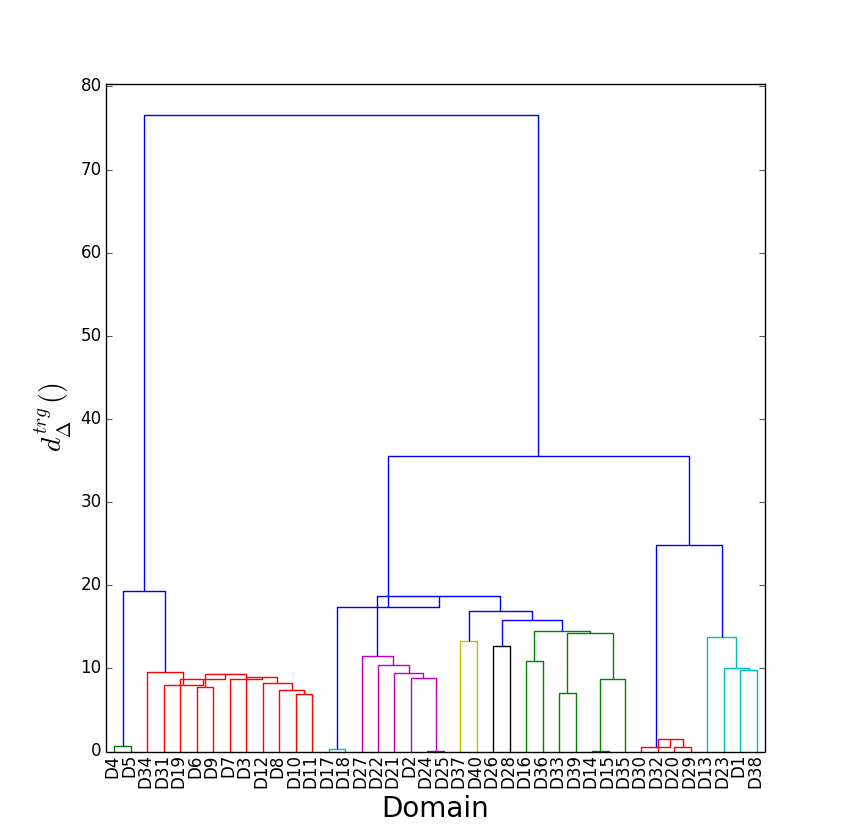}}
\caption{Dendrograms of the HAC algorithm instantiated with: (a) ${\tt d}^{\tt trg}_{\tt PP}()$ and (b) ${\tt d}^{\tt trg}_{\Delta}()$.}
\label{fig:dnd:pp}
\end{figure*}

\begin{figure*}[ht]
\centering
 \subfloat[]{\includegraphics[width=0.5\textwidth]{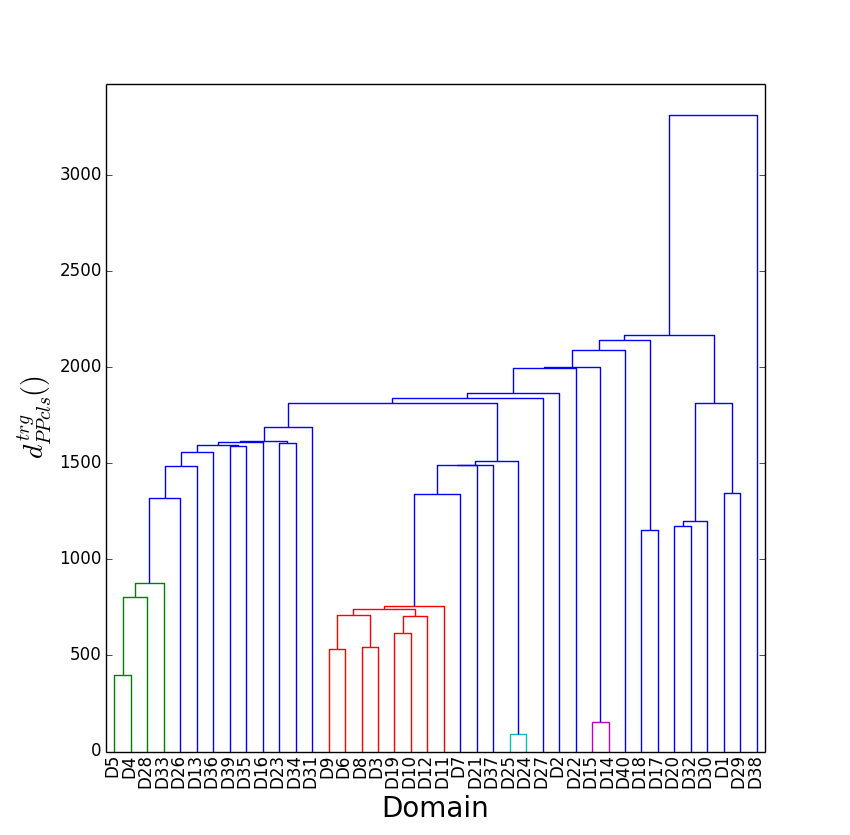}}
  \subfloat[]{\includegraphics[width=0.5\textwidth]{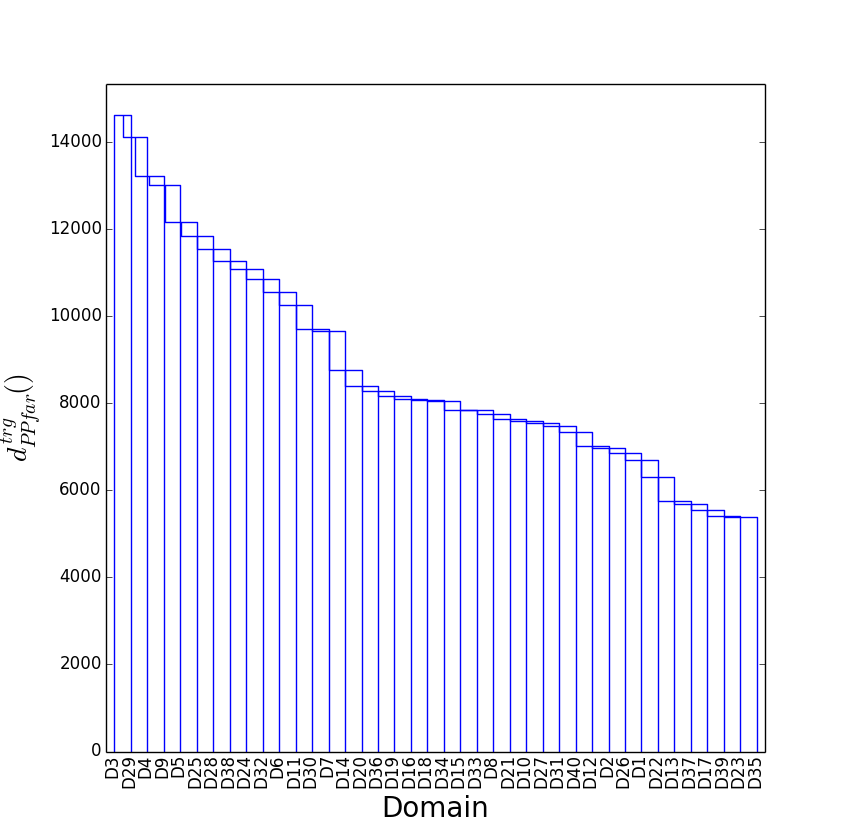}}
\caption{Dendrograms of the HAC algorithm instantiated with: (a) ${\tt d^{\tt trg}_{PPcls}}()$ and (b) ${\tt d^{\tt trg}_{PPfar}}()$.}
\label{fig:dnd:singlelink}
\end{figure*}

Figures~\ref{fig:dnd:pp} and~\ref{fig:dnd:singlelink} show the
dendrograms relative to four (out of five) distances. Each horizontal
line segment indicates the merging of two clusters; the length of the
vertical line segments incident to the extremes of the horizontal line
segment is proportional to the distance between the merged cluster and
each cluster to merge: the shorter, the more convenient the merging;
viceversa, the longer, the better to avoid the merging. With this key,
the dendrograms can be read as follows: ${\tt d^{\tt trg}_{PPcls}}()$,
${\tt d}^{\tt trg}_{\tt PP}()$ and ${\tt d}^{\tt trg}_{\Delta}()$ show
a growing, but anyway interesting, ability in grouping original TMs in
few, compact clusters. It is a matter of fact that such ability
increases with the computational cost (cf. Table~\ref{tab:proctime}).
On the contrary, ${\tt d^{\tt trg}_{PPfar}}()$ does not work at all:
the well known {\em chaining} effect is here observed, likely because
the low number of points to cluster. A similar behaviour is seen with
${\tt d}^{\tt src}_{\tt CA}()$, for which we then omit the dendrogram.

\subsection{Intrinsic evaluation: Silhouette}
\label{sec:intrEval}

\begin{figure}[ht]
\centering
    \includegraphics[width=0.5\textwidth]{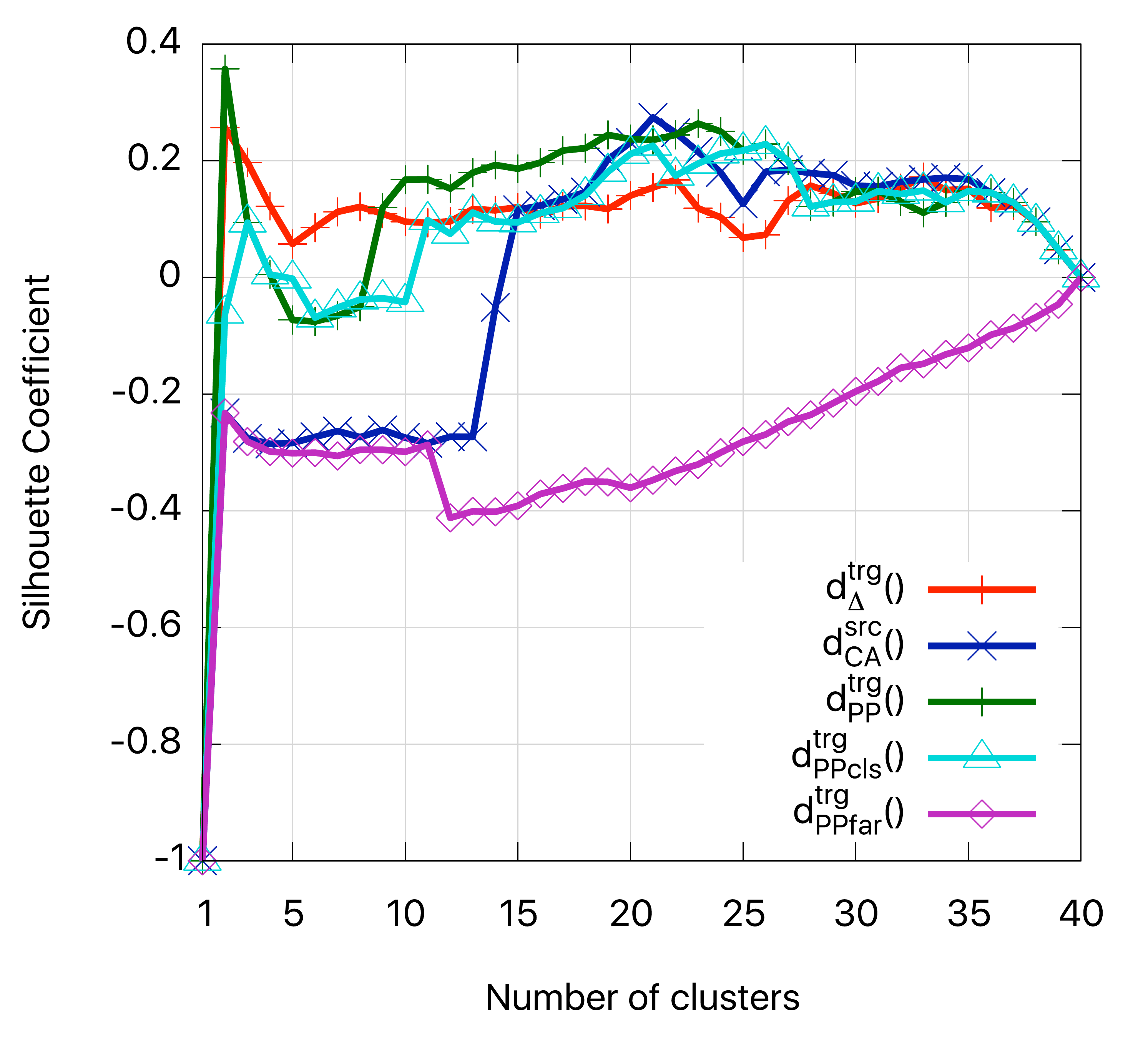}%
 \caption{Silhouette curves of the five instances of the HAC algorithm.}
\label{fig:SC}
\end{figure}

As discussed in Section~\ref{sec:evaluation}, different clusterings
generated during the run of a clustering algorithm or by different
clustering algorithms can be intrinsically evaluated through the
Silhouette value. Figure~\ref{fig:SC} plots the Silhouette for the
clusterings generated at each iteration by our HAC algorithm
instantiated with the five proposed distances.  The number of clusters
in each clustering is reported on the abscissa; hence, in order to see
how the algorithm proceeds from the first to the last iteration, the
plot should be seen right-to-left.

First of all, we observe that the values are quite low, exceeding
rarely even 0.25, not a really high value. This is due to the
co-occurrence on the one hand of the low number of points to cluster
(40) and on the other of the arbitrary setting to 0 of the Silhouette
value for the single-point clusters (Section~\ref{sec:evaluation}). In
fact, in early HAC iterations there are many 0-valued clusters -- fact
that lowers the overall Silhouette of clusterings -- that disappear
altogether only at the cost of straggly clusters -- which still keep
the coefficient low.

Apart that, the plot confirms the ineffectiveness of ${\tt d^{\tt
    trg}_{PPfar}}()$ predicted by the corresponding dendrogram. The
other distances show similar values for the first 10-15 iterations
(i.e. clusterings with 25-30 to 40 clusters); after that, they start
to diversify: ${\tt d^{\tt trg}_{PP}}()$, ${\tt d^{\tt
    trg}_{PPcls}}()$ and ${\tt d^{\tt src}_{CA}}()$ initially generate
good clusterings that tend to gradually worsen with further
aggregations, until an abrupt drop occurs. On the contrary, ${\tt
  d^{\tt trg}_{\Delta}}()$ keeps the same coefficient even with very
few clusters. The peaks reached at the last but one iteration
(clusterings with just two clusters) derive from the disappearance of
single-point, i.e. 0-valued, clusters.

\subsection{Extrinsic evaluation: BLEU and PP}

Clusterings can also be indirectly compared by looking at performance
of tasks where they are used; this is the so called extrinsic
evaluation (Section~\ref{sec:evaluation}). Since here clusterings are
used for inducing a decomposition of SMT models in domains, two
straightforward extrinsic evaluations are the perplexity of the
induced LMs and the final MT quality measured in terms, for example,
of BLEU score.

\begin{figure}[ht]
\centering
\includegraphics[width=1.1\columnwidth,height=70mm]{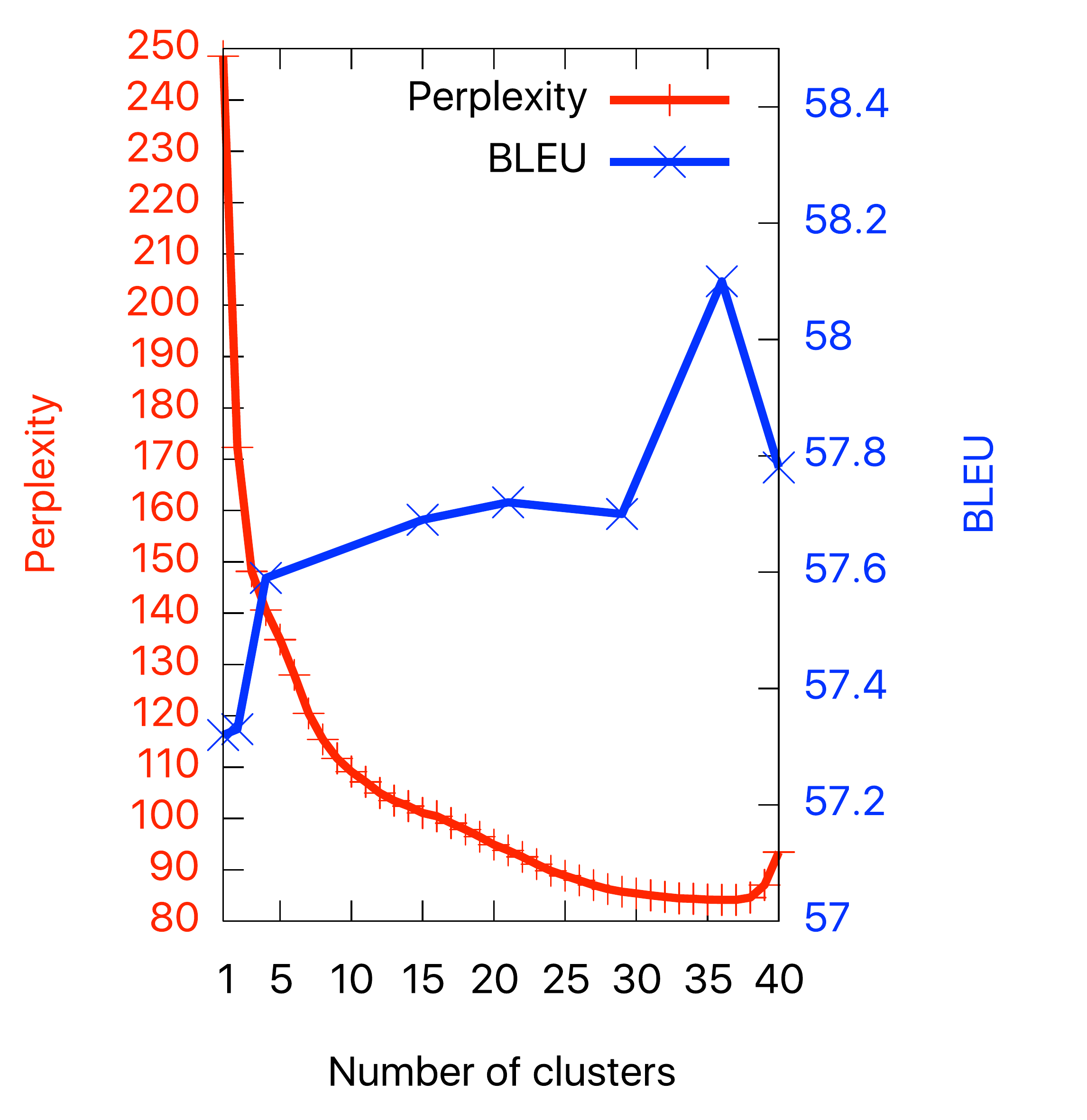}
\caption{PP curve and BLEU scores of the HAC algorithm instantiated with ${\tt d}^{\tt trg}_{\Delta}()$.}
\label{fig:pp-bleu:d_delta}
\end{figure}

\begin{figure}[ht]
\centering
\includegraphics[width=1.1\columnwidth,height=70mm]{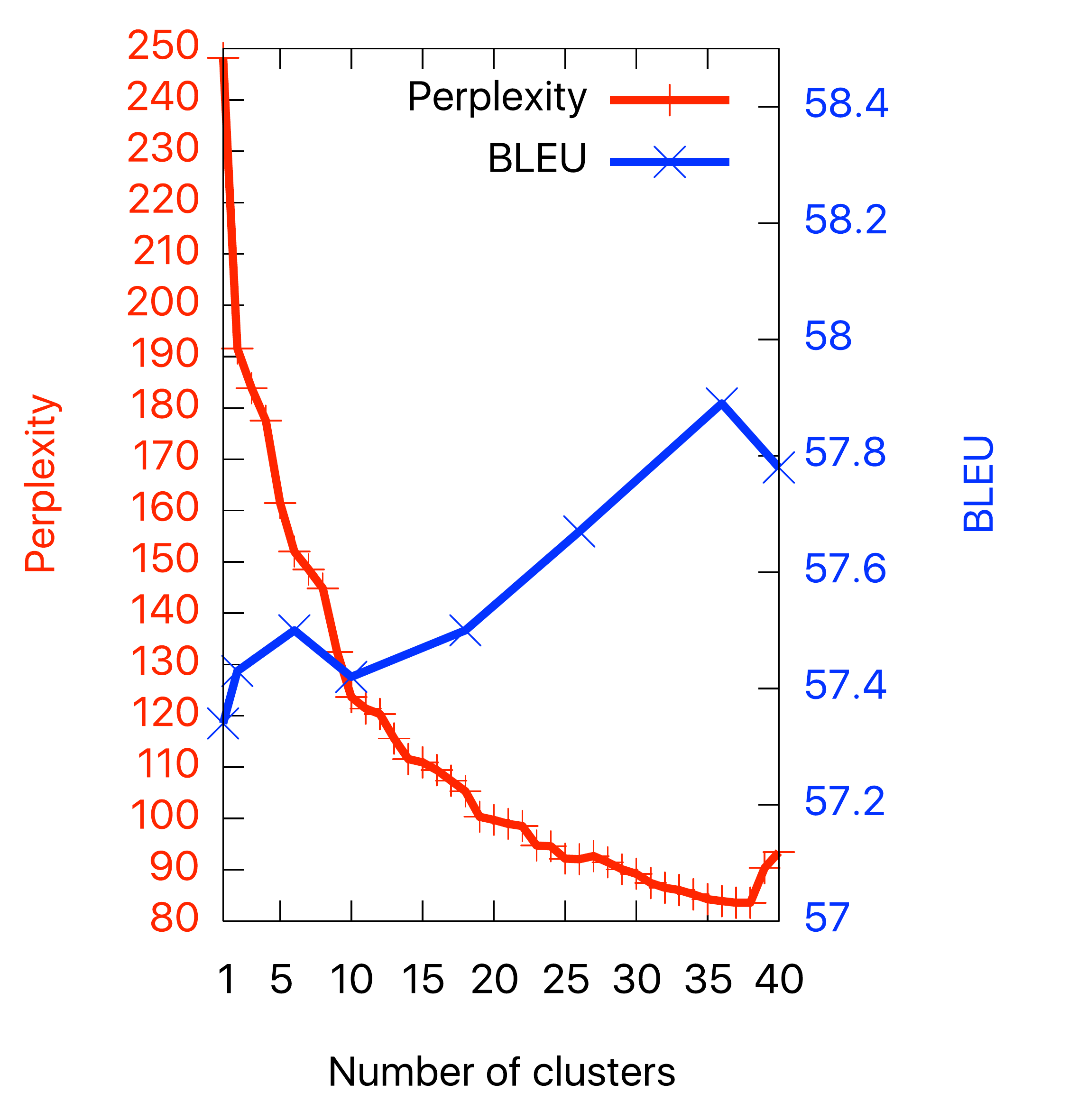}
\caption{PP curve and BLEU scores of the HAC algorithm instantiated with ${\tt d}^{\tt trg}_{\tt PP}()$.}
\label{fig:pp-bleu:d_pp}
\end{figure}

Figures~\ref{fig:pp-bleu:d_delta} and~\ref{fig:pp-bleu:d_pp} plot the
perplexity and the BLEU score of the clusterings generated during the
40 iterations of our algorithm instantiated with the two most
promising distances, according to the dendrograms, that is ${\tt
  d}^{\tt trg}_{\Delta}()$ and ${\tt d}^{\tt trg}_{\tt PP}()$.

The values measured in correspondence of the two extreme clusterings
(at the beginning when each domain is a cluster in its own and at the
end when all domains have been agglomerated into one single cluster)
are of course equal whatever the instance of the HAC algorithm.

Concerning the perplexity, the two distances behave very similarly,
${\tt d}^{\tt trg}_{\Delta}()$ being a bit smoother than ${\tt d}^{\tt
  trg}_{\tt PP}()$ thanks to the possibility to choose the best local
merging in a more reliable way. In particular, a slight improvement is
observed after early aggregations; successively, a gradual degradation
occurs which becomes more severe in the last 10 iterations. It is
worth to note that the perplexity trend resembles quite closely that of
the Silhouette coefficients of Figure~\ref{fig:SC}, apart the outlier
peaks of the latter in correspondence of clusterings with 2 clusters.

Also the BLEU curves are quite well predicted by both the Silhouette
and the perplexity: a tiny improvement at the beginning; then, a
plateau for ${\tt d}^{\tt trg}_{\Delta}()$ and a slight degradation
for ${\tt d}^{\tt trg}_{\tt PP}()$; finally, a rather sharp fall for
both distances.

We can now answer the two questions posed in the introduction. In
fact, as shown, ${\tt d}^{\tt trg}_{\Delta}()$ and ${\tt d}^{\tt
  trg}_{\tt PP}()$ allow to improve the BLEU score of the original
domain-specific models by merging few, very close domains (question
ii), while, more importantly, ${\tt d}^{\tt trg}_{\Delta}()$ is even
able to keep the degradation of the BLEU score under 0.2 absolute
points (from 57.8 to 57.6) employing just 5 specialised models instead
of 40 (question i).

On the other side, it should be said that the BLUE score does not vary
too much, being the difference between the highest and the lowest
values lesser than 1 absolute point; this calls for an assessment on
a more challenging benchmark.

\subsection{Discussion}

Often we read that ``clustering is an art, not a science'' and that
choosing the right way to measure the distance between the points of
the task at hand is even more important than the clustering algorithm
actually employed. Those remarks are confirmed by our investigation.
The same HAC algorithm was instantiated with five distances and its
behaviour observed from different points of view: dendrograms,
intrinsic and extrinsic evaluations.

The outcomes of such views are different: for example, according to
Silhouette, ${\tt d}^{\tt src}_{CA}()$ is effective while its
dendrogram is very bad; again, the perplexity curves of ${\tt d}^{\tt
  trg}_{\Delta}()$ and ${\tt d}^{\tt trg}_{\tt PP}()$ are practically
indistinguishable, while the dendrogram of the former appears to be
better than the dendrogram of the latter.

Each single view can also be affected by critical aspects that should
be taken into account. For example, in our particular set-up, the
Silhouette coefficient is highly affected by the 0-valued clusters,
while the dendrograms by having taken the entire TMs as atomic points
to aggregate.

Hence, only an overall view of all measures can suggest reliable
conclusions; and our measures, as a whole, suggest that ${\tt d}^{\tt
  trg}_{\Delta}()$ is the most effective distance out of those tested.

\section{Summary and Future Work}
\label{sec:conclusion}

In this paper we have summarised our investigation on domain
clustering in the ambit of an adaptive MT architecture. A standard
bottom-up hierarchical clustering algorithm has been instantiated with
five different distances, which have been compared, on an MT benchmark
with 40 commercial domains, in terms of dendrograms, intrinsic and
extrinsic evaluations.  The main outcome is that the most expensive
distance is also the only one which allows the MT engine with just few
cluster-specific models to perform as well as the 40-domains adapted MT
engine.

In the close future, we are going to extend the here reported
investigation as follows. First of all, instead of considering each
original TM as an indivisible, single point, a finer granularity will
be considered to both overcome the 0-valued clusters issue
(Section~\ref{sec:intrEval}) and improve the performance of
single-link instances of the HAC algorithm. Unfortunately, no further
meta-information is provided inside our TMs in addition to the
identity of the customer who provided it. Anyway, finer
straightforward single points to aggregate could be: (i)~single
segments inside TMs; (ii)~automatic clusters of sentences inside each
TM. 

Second, our evaluations treated equally all words, but a customer
could consider more valued the proper translation of domain-specific
terminology than of other words. For this reason, we are manually
annotating domain specific terms in Benchmark 1.1 for comparing the
instances of the HAC algorithm with respect to them.

Finally, we will test the clustering on much more challenging
benchmarks with hundred to even thousand domains.

\section*{Acknowledgments}
FBK authors were supported by the MMT project which received funding from the EU's Horizon 2020 research and innovation programme under grant agreement No 645487.

\bibliographystyle{eacl2017}

\end{document}